\documentclass{bmvc2k}

\usepackage{enumitem}   %,ulem}    %% remove ulem once paper is finished and ready to submit
\usepackage{amsfonts}
\usepackage{footnote}

\graphicspath{ {img/} }

\title{Semantically Selective Augmentation for Deep Compact Person Re-Identification}

\addauthor{V\'ictor Ponce-L\'opez$\ast$ $^{1}$}{http://victorponce.info}{1}
\addauthor{Tilo Burghardt $^{1}$}{http://people.cs.bris.ac.uk/~burghard}{1}
\addauthor{Sion Hannunna $^{1}$}{}{1}
\vspace{-1cm}
\addauthor{Dima Damen $^{1}$}{http://people.cs.bris.ac.uk/~damen/}{1}
\addauthor{Alessandro~Masullo $^{1}$}{}{1}
\vspace{-5mm}
\addauthor{Majid Mirmehdi $^{1}$}{http://people.cs.bris.ac.uk/~majid/}{1}

\addinstitution{
 $^{1}$ Dept. of Computer Science\\
 University of Bristol\\
 Merchant Venturers Building\\
 Woodland Road, BS8 1UB, UK\\
 \vspace{28mm}\scriptsize $\ast$Correspondence: \emph{vponcelop@gmail.com}\\ 
 \scriptsize \emph{\{v.poncelopez, tb2935, sh1670, dima.damen, a.masullo, m.mirmehdi\}@bristol.ac.uk}
}

\runninghead{Ponce-L\'opez \bmvaEtAl}{Semantically Selective Augmentation for Re-ID}

\def\eg{\emph{e.g}\bmvaOneDot}

\def\ie{\emph{i.e}\bmvaOneDot}

\def\wrt{{\emph{w.r.t}\bmvaOneDot} }

\begin{document}

\maketitle

%\vspace{-2mm}
\begin{abstract}
{We present a deep person re-identification approach that combines semantically selective, deep data augmentation  with clustering-based network compression to generate high performance, light and fast inference networks. In particular, we propose to augment  limited training data via sampling from a deep convolutional generative adversarial network (DCGAN), whose  discriminator is constrained by a semantic classifier to explicitly control the domain specificity of  the generation process. Thereby, we encode information  in the  classifier network which  can be utilized to steer  adversarial synthesis, and which fuels our CondenseNet ID-network training. We provide a quantitative and qualitative analysis of the approach and its variants on a number of datasets, obtaining results that outperform the state-of-the-art on the LIMA dataset for long-term monitoring in indoor living~spaces. }
\end{abstract}

%-------------------------------------------------------------------------
%\vspace{-3mm}
\section{Introduction}
\label{sec:intro}
%\vspace{-1mm}

%\textcolor{green}{
%\mmn{Victor please find top-class references for those marked as [?] below. \\}
Person re-identification (Re-ID) across cameras with disjoint fields of view, given unobserved intervals and varying appearance (\eg change in clothing), remains a challenging subdomain of computer vision. The task is particularly demanding whenever facial biometrics~\cite{Yu_2016_CVPR} are not explicitly applicable, be that due to very low resolution~\cite{face_lowRes2017} or non-frontal shots. Deep learning approaches have recently been customized, moving the domain of person Re-ID forward~\cite{BARBOSA201850} with potential impact on a wide range of applications, for example, CCTV surveillance~\cite{deep-surveillance_2016} and e-health applications for living and working environments~\cite{Sadri_AmI}. Yet, obtaining cross-referenced ground truth over long term~\cite{mcconville2018understanding,twomey2016sphere},  realising deployment of inexpensive inference platforms,  and establishing visual identities from strongly limited data -- all remain fundamental challenges. In particular, the dependency of most deep learning paradigms on vast training data pools and high computational  requirements for heavy inference networks appear as significant challenges to many person Re-ID~settings.

\begin{figure}
     \centering
     %\vspace{-3mm}
     \includegraphics[width=\linewidth,height=60mm]{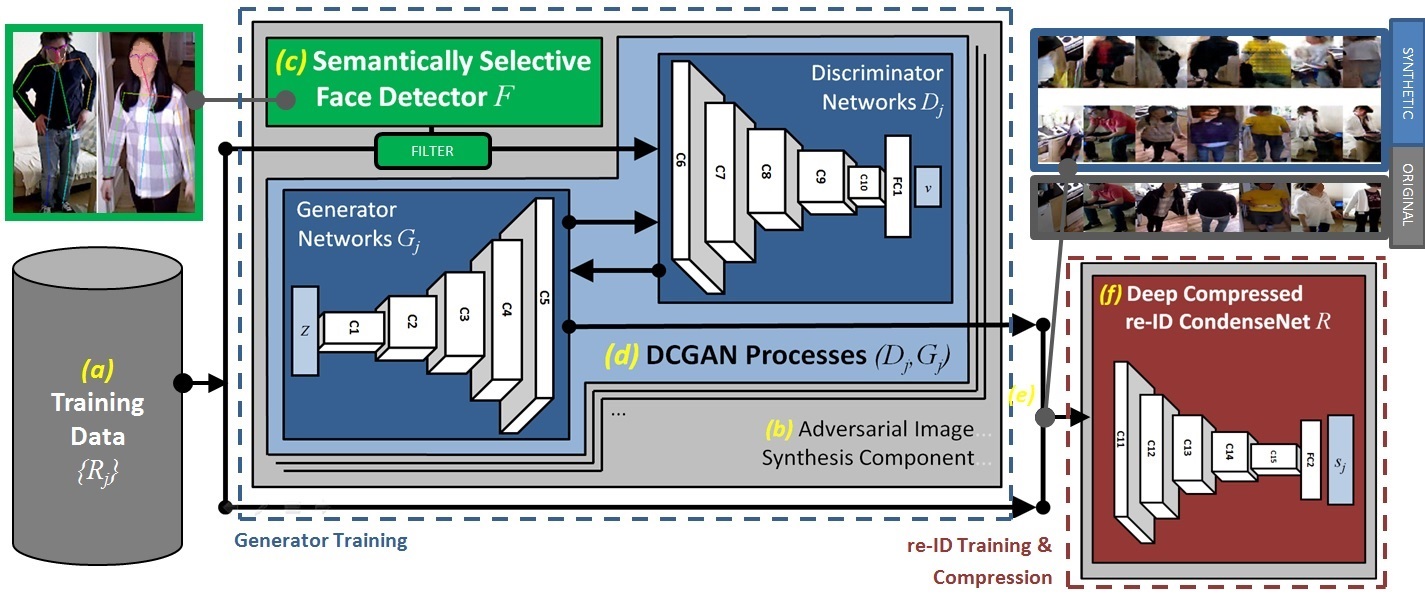}
     %\vspace{-7mm}
     \caption{{\small{\textbf{Framework Overview.} Visual deep learning pipeline at the core of our approach: inputs~(dark gray) are semantically filtered via a face detector (green) to enhance adversarial augmentation via DCGANs~(blue). Original and synthetic data is  combined to train a compressed CondenseNet~(red) producing a light and fast ID-inference network.} } }
     % should R_j in the figure be X_j for the training data? I understand this refer to the CondenseNet inputs, however it is not defined anywhere and I think it may be clearer.
     \label{fig:overview}
     %\vspace{-2mm}
\end{figure}

%\textcolor{green}
{In this paper, we introduce an approach for producing high performance, light and fast deep Re-ID inference networks for persons - built from limited training data and not explicitly dependent on face identification. To achieve this, we propose an interplay of three recent deep learning technologies as depicted in Figure~\ref{fig:overview}:  deep convolutional adversarial networks (DCGANs)~\cite{RadfordMC15} as class-specific sample generators (in blue); face detectors~\cite{simon2017hand} used as semantic guarantors to steer synthesis (in green);  and a clustering-based CondenseNet~\cite{huang2017condensenet} as a compressor (in red). We  show that the proposed face-selective adversarial synthesis allows to generate new, semantically selective and meaningful artificial images that can improve subsequent training of compressive ID networks. Whilst the training cost of our approach can be significant due to the adversarial networks' slow and complicated convergence {process~\cite{NIPS2014_5423-GAN}}, our parameter count of final CondenseNets is approximately one order of magnitude smaller than those of other state-of-the-art systems, such as ResNet50{~\cite{zheng2017unlabeled}}. We provide a quantitative and qualitative analysis {over different adversarial synthesis paradigms for our} approach, obtaining results that outperform the highest achieving published work on the LIMA dataset \cite{LayneLIMA17}  for long-term monitoring in indoor living environments. First, we will provide a brief overview of works related to the proposed approach.}

%\vspace{-3mm}
\section{Related Work}
\label{sec:sota}
%\vspace{-1mm}

Technologies applicable to perform person Re-ID form a large and long-standing research area with considerable history~and specific associated challenges~\cite{zheng2016}. Whilst low-resolution face recognition~\cite{face_lowRes2017}, gait and behaviour\ analysis~\cite{Takemura2018}, as well as full-person, appearance-based recognition~\cite{zheng2016} all offer routes to performing `in-effect' person ID or Re-ID, for this brief review we will focus on particular technical aspects, \ie looking specifically at recent  augmentation and deep learning approaches for appearance-based methods.   

\textbf{Augmentation - } Despite improvements in {methods for high-quality, high-volume} ground truth acquisition~\cite{mcconville2018understanding,ThanWifivisual2017}, {input data augmentation~\cite{Perez2017TheEO}} remains a key strategy to support generalisation in deep network training generally. It allows to expose networks to otherwise inaccessible pattern configurations to back-propagate against, which, if representing realistic and relevant content, improves generalisation potential of the training procedure.
The use of synthetic data in the training set presents several advantages, such as the ability to reduce the effort of labeling images and to generate customizable domain-specific data. It has been noted that combining synthetic and measured input often shows improved performance over using synthetic images only~\cite{ShrivastavaPTSW16}. 
 Recent examples of non-augmented, innovative approaches in the person Re-ID  domain  include feature selection strategies \cite{Long-term_reid,surveillance_2017}, antropometric profiling~\cite{Bondi2017LongTP} using depth cameras, and multi-modal tracking~\cite{ThanWifivisual2017}, amongst many others. Augmentation has long been used in Re-ID scenarios too, for instance in~\cite{BARBOSA201850}, the authors consider the structural aspects of the human body by exploiting mere RGB data to fully generate semi-realistic synthetic data as inputs to train neural networks, obtaining promising results for person Re-ID. Image augmentation techniques have also demonstrated their effectiveness in improving the discriminative ability of learned CNN embeddings for person Re-ID, especially on large-scale datasets~\cite{zheng2017unlabeled,BARBOSA201850,Chen_DeepMulti}. Recently, the learning and construction of the modelling space {itself}, used for augmentation, has been realised in deep adversarial learning  architectures.

\textbf{Adversarial Synthesis - } Generative Adversarial Networks~(GANs)~\cite{NIPS2014_5423-GAN} in particular have been widely and successfully applied to deliver augmentation -- mainly building on their ability to construct a latent space that underpins the training data, and to sample from it to produce new training information.  DCGANs~\cite{RadfordMC15} pair the GAN concept with compact convolutional operations to synthesise visual content more efficiently. The DCGAN's ability to organise the relationship between a latent space and an actual image space associated to the GAN input has been shown in a wide variety of applications, including face and pose analysis~\cite{RadfordMC15,NIPS2017_6644}. In these and other domains, latent spaces have been constructed that can convincingly model and parameterise object attributes such as scale, rotation, and position from unsupervised models, and hence dramatically reduce the amount of data needed for conditional generative modeling of complex image distributions.

\textbf{Compression and Framework - } Given ever-growing computational  requirements for very-deep inference networks, recent research into network compression and optimisation has produced a number of approaches capable of compactly capturing network functionality. Some examples include ShuffleNet~\cite{ZhangZLS17ShuffleNet}, MobileNet~\cite{HowardZCKWWAA17MobileNet}, and CondenseNet~\cite{huang2017condensenet}, which have proven to be effective even when operating on small devices where computational resources are limited. 

Here, we combine semantic data selection for data steering, adversarial synthesis for training space expansion, and CondenseNet compression to sparsify the built Re-ID classifier representation. Our solution operates on single images during inference, able to perform the Re-ID step in a one-shot paradigm\footnote{Whilst results are competitive in this setting, discovering and matching segments during inference~\cite{LayneLIMA17,LIU2018650,ZHOU2018739,WU2018727,BMVC2015_129} is not used and could potentially further improve performance.}. The following section details the components and functionality of the proposed pipeline.

%\vspace{-3mm}
\section{Methodology and Framework Overview}
%\vspace{-1mm}
{Figure~\ref{fig:overview} illustrates our methodology pipeline, which follows  a generative-discriminative paradigm: \textbf{(a)}  training data sets~$\{X_j\}$ of image patches are produced by a person detector, where each image patch set is either associated to a known person identity label~$j\in{\{1,..,N\}}$, or an `unknown' identity label~$j=0$. \textbf{(b)}~An image augmentation component  then expands on this dataset. This component consists of \textbf{(c)} a facial filter network~$F$ based on multi-view bootstrapping { and OpenPose}~\cite{simon2017hand}; and \textbf{(d)} DCGAN~\cite{RadfordMC15} processes, whose  discriminator networks~$D_j$ are  constrained by the semantic selector~$F$ to control domain specificity. The set of DCGANs, namely network pairs~$(D_j,G_j)$, are employed to train  generator networks~$G_j$ that synthesise unseen samples~$x$ associated with labels~$j\in\{0,..,N\}$. These generators~$G_j$ are then used to produce large sets of samples. We focus on two types of scenarios: \textbf{(1) }a setup where we synthesize content for each identity class~$j$ individually, and \textbf{(2)} one where only a single `unlabeled person' generator $G$ is produced using all classes~$\{X_j\}$ as input, with the aim to generate generic identity content, rather than individual-specific imagery. Sampled output from generators is \textbf{(e)} unified with the original frame sets and labels, forming the input data for \textbf{(f)} training a {Re-ID} CondenseNet~$R$ that learns to map sample image patches~$x_j$ to ID score vectors~$s_j\in\mathbb{R}_{+}^{(N+1)}$ over all identity classes. This yields the sparse inference network~$R$ built implicitly compressed in order to support lightweight inference and deployment via a single network. Each component is now considered in detail.}

%\vspace{-3mm}
\subsection{Adversarial Synthesis of Training Information} 
\label{sec:dcgan}
%\vspace{-1mm}
\hspace{5mm}{\textbf{Adversarial Network Setup -} We utilise the generic adversarial training process of DCGANs~\cite{RadfordMC15} and its suggested network design in order to construct a de-convolutional, generative function~$G_j$ per synthesised label class $j\in\{0,..,N\}$ that -- after training -- can produce new images~$x$ by sampling from a sparse latent space~$Z$. Depending on the experiment, a single `generic person' network $G$ may be built instead utilising all $\{X_j\}$. 
As in all adversarial setups, generative networks~$G$ or $\{G_j\}$ are paired with discriminative  networks~$D$ or $\{D_j\}$, respectively. The latter map from images~$x$ to an `is synthetic' score~$(v=D(x))>0$, reflecting network support for~$x\notin \{X_j\}$. Essentially, the discriminative networks then learn to differentiate generator-produced patches ($v$\textgreater\textgreater) from original patches~($v$<<). However, we add to this classic dual network setup~\cite{NIPS2017_6644}, a third externally trained classifier~$F$ that filters and thereby controls/selects the input to~$D_j$ -- in our case that is restricting input to those samples where the presence of faces can be established\footnote{We also modify the initial layer of the DCGAN to deal with a temporal gap of the specified number of frames. %{Relevant source code will be made available} 
\href{https://github.com/vponcelo/DCGAN-tensorflow/}{https://github.com/vponcelo/DCGAN-tensorflow}}.
}
% For simplicity, I think it is clearer that we denote $x$ for the generated images iff we do not need to refer to its associated original class (as I think we don't), and $x_j$ to the original sample image of class j where $x_j\in \{Xj\}$ as we do need in the Training Process below

{\textbf{Facial Filtering -} We use the face keypoint detector from OpenPose~\cite{simon2017hand} as the filter network~$F$ to semantically constrain  the input to~$D_j$ and $D$. This method utilises multi-view boostrapping applied to face detection: it maps from images to facial keypoint detections, each with an associated detection confidence. If at least one such keypoint can be established then  face detection is defined as successful, where formally~$F( x_j\in X_j)\in{}[0,1]$ is assigned to reflect either the absence~$(0)$ or presence~$(1)$ of a face. 
}
\begin{figure}
     \centering
     %\vspace{-1mm}
     \includegraphics[width=.95\linewidth]{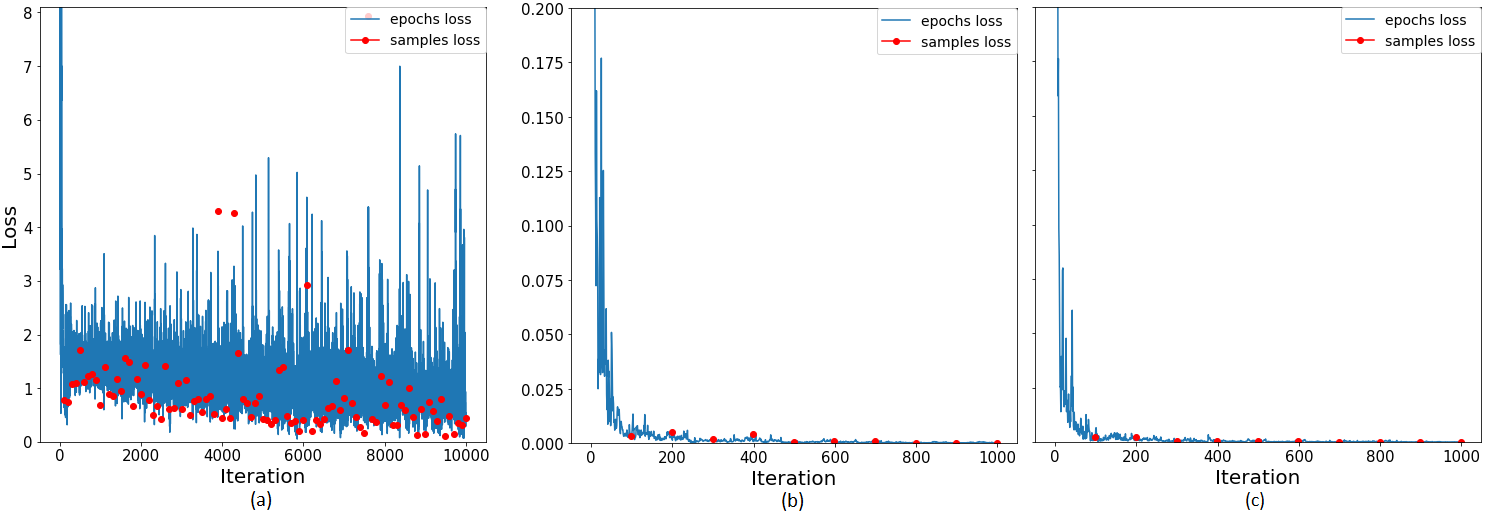}
     %\vspace{-4mm}
     \caption{\small \textbf{DCGAN Training.} \textbf{(a)} development of loss while training $(D,G)$ as a base network using all $\{X_j\}$; \textbf{(b)} reduced loss and fast convergence when re-training to obtain a new pair $(D,G)\Rightarrow(D_0, G_0)$ from a set $X_0$ (`Unknown' identity); and \textbf{(c)} re-training to obtain a new pair $(D,G)\Rightarrow(D_1, G_1)$ with the semantic controller~$F$. Discriminator losses are shown for $D$ (pre-trained) and $D_0$, $D_1$ (re-trained) densely for original samples (blue) and sparsely every 100 iterations for generated samples (red).}
     \label{fig:dcgan-training}
     %\vspace{-2mm}
\end{figure}

{\textbf{Training Process - } All networks  then engage in an adversarial training process utilising Adam~\cite{Kingma2014AdamAM} to optimise the networks $D$, $\{D_j\}$, and $G$, $\{G_j\}$, respectively, according to the discussion in~\cite{RadfordMC15}, whilst enforcing the domain semantics via~$F$. The following detailed process describes this training regime: \textbf{(1)} each~$D$ or $D_j$ is optimised towards minimising the negative log-likelihood~$-log(D(x))$ based on the relevant inputs from $\{X_j\}$ iff $F(x_j)=1$, \ie on original samples that are found to contain faces.  \textbf{(2)}~Network optimisation then switches to back-propagating errors into the entire networks $D(G(z))$ or $D_j(G_j(z))$, respectively, where~$z$ is sampled from a randomly initialised Gaussian to generate synthetic content. Consider that whilst the generator weights are adjusted to minimise the negative log-likelihood~$-log(D(G(z)))$, encouraging $v$ to get lower scores, the discriminator weights are adjusted to maximise it, prompting $v$ to get higher scores. DCGAN training then proceeds by alternating between \textbf{(1)} and \textbf{(2)} until acceptable convergence.
}

%\textcolor{green}{

{\textbf{Intuition behind Semantically Selective Adversarial Training -} Consider that training proceeds by, for instance, optimising~$G_j$ to produce synthetic images of a kind  that~$D_j$ cannot differentiate from face-containing samples in~$X_j$ using $F$ as a semantic guarantor. Concurrently,~$D_j$ is trained to differentiate images produced by~$G_j(z)$ from original face-containing samples in~$X_j$. These two processes are antagonistic -- they cannot both perfectly achieve their optimisation target, and instead will approach a Nash equilibrium in a successful training run~\cite{RadfordMC15}. As a result, the properties of samples from the set~$X_j$ and the ones generated by~$G_j(z)$ will move towards convergence, without~$G_j$ being restricted to the original sample set.  This aims at constructively generalising the information content captured by within original input. The same rationale is of course applicable to images produced by $G(z)$, again with $F$ as the semantic guarantor for face-content, but this time aiming at the synthesis of generic person imagery rather than individual-specific content. Note that the network $G$ -- once trained -- can thus also serve as a suitable `pre-trained' basis network for optimising individual-specific generators $\{G_j\}$ faster (see Figure~\ref{fig:dcgan-training}). 
} 
%\textcolor{red} {(IMPORTANT SUGGESTION:  technically rigorous reviewers will pick up on that and it could well improve results -- in order to fully explore the rationale of semantic filtering it would be required to also experiment with enforcing $F(G_j(z))=1$ as a person detector DURING adversarial training, that is overriding $D_j$ and forcefully setting the cost to maximal if the former is not the case in every iteration of (2)! This alters the cost function of $D_j(G_j(z))$ to one that punishes~$G_j$ for image generation that does not contain faces instead of just picking more face images in $X_j$. 
%ADDED SUGGESTION (Fig 2): May I also suggest, that pre-training all generators with all face-containing images before adversarial training as well as unifying the DCGAN to a single network for all classes are potentially  important next research steps/experiments --> .)}  % Running experiment for RED. if not best results obtained, then just delete red.

%\vspace{-3mm}
\subsection{Re-ID Network Training and Compression }
%\vspace{-1mm}

Once the synthesis networks~$G$ and $\{G_j\}$ are trained, we sample their output and combine it with all original training images (withholding 15\% per class for testing) to train $R_{}$ as a CondenseNet~\cite{huang2017condensenet}, optimised via standard stochastic gradient decent with Nestrov momentum. Structurally, $R_{}$ maps from $256\times 256$-sized RGB-tensors to a score vector over all identity classes. We perform 120 epochs of training on all layers, where layer-internal grouping is applied to the dense layers in order to actively structure network pathways by means of clustering~\cite{huang2017condensenet}. This principle has been proven effective in DenseNets~\cite{huang2017densely}, ShuffleNets~\cite{ZhangZLS17ShuffleNet}, and MobileNets~\cite{HowardZCKWWAA17MobileNet}. However, CondenseNets extend this approach by introducing a compression mechanism to remove low-impact connections by discarding unused weights. As a consequence, the  approach produces an ID inference network\footnote{\href{https://github.com/vponcelo/CondenseNet/}{https://github.com/vponcelo/CondenseNet/}} which is implicitly compressed and supports lightweight deployment.

\begin{figure}[t]
     \centering
     \includegraphics[width=\linewidth]{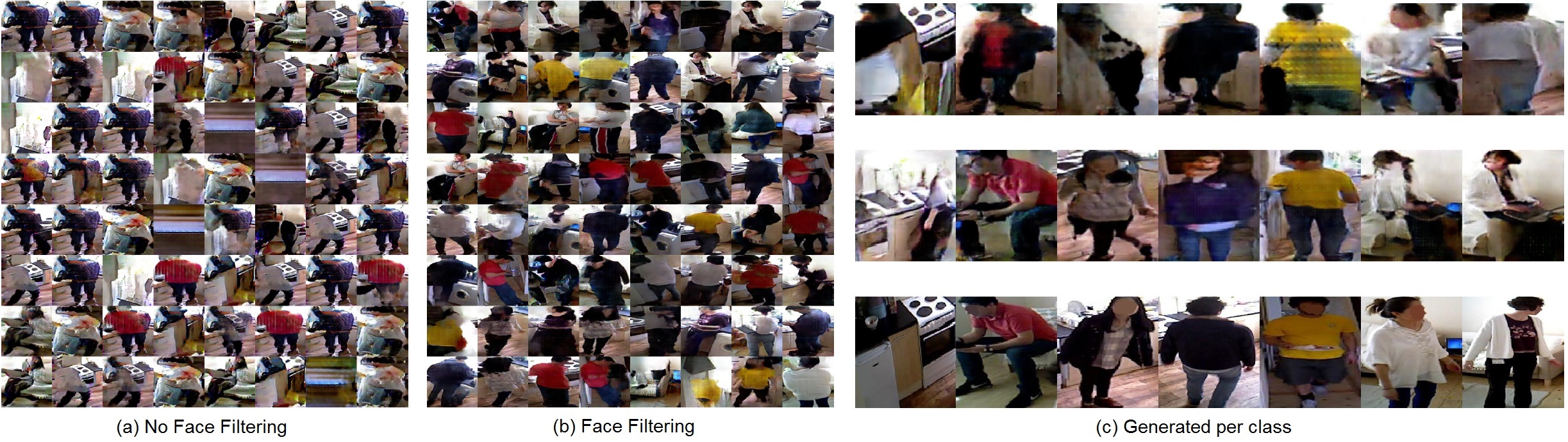}
     %\vspace{-5mm}
     \caption{\small{ \textbf{DCGAN Synthesis Examples.} Samples generated by  $G(z)$ with \textbf{(b)} or without \textbf{(a)} semantic controller.  (\textbf{c}) $1^{st}$ row: Examples of generated images from $G_0$ and $G_j$ without semantic controller; $2^{nd}$ row: with semantic controller;  $3^{rd}$ row: original samples from $X_0$ and $\{X_j\}$. Columns in (\textbf{c}) are, from left to right, `unknown' identity $0$ and identities $j\in \{1,...,N\}$, respectively.}}
     %\vspace{-3mm}
     \label{fig:DCGAN_samples}
\end{figure}

%\vspace{-5mm}
\subsection{Datasets }
\label{sec:exp}

\hspace{5mm}\textbf{DukeMTMC-reID - }
First we confirm the viability of a GAN-driven CondenseNet  application in a traditional Re-ID setting (\eg larger cardinality of identities, outdoor scenes) via the DukeMTMC-reID~\cite{DukeMTMC-reID} dataset, which  is a subset of a multi-target, multi-camera pedestrian data corpus. It contains eight 85-minute high-res videos with pedestrian bounding boxes. It covers $1,812$ identities, where $1,404$ identities appear in more than two cameras and $408$~identities (distractor IDs)  appear in only one\footnote{More details about the evaluation protocol in \href{https://github.com/layumi/DukeMTMC-reID_evaluation}{https://github.com/layumi/DukeMTMC-reID\_evaluation}.\label{eval-reid}}. 

\textbf{Market1501 - } We also use a large-scale person Re-ID dataset called Market1501~\cite{BowKissme15} collected from 6 cameras covering $1,501$ different identities across $19,732$ images for testing and $12,936$ images for training generated by a deformable part model (DPM)~\cite{DPM10}. 

\textbf{LIMA - } The \textbf{L}ong term \textbf{I}dentity aware \textbf{M}ulti-target multi-camer\textbf{A} tracking dataset \cite{LayneLIMA17}, provides us with our main test bed for the approach. In contrast to previous datasets, image resolution is high enough in this dataset to effectively apply face detection as a semantic steer. LIMA contains a large set of $188,427$ images of identity-tagged bounding boxes gathered over 13 independent sessions, where bounding boxes are estimated based on OpenNI NiTE %\footnote{http://openni.ru/files/nite/} 
operating on RGB-D and are grouped into time-stamped, local tracklets. The dataset covers a small set of $6$ individuals filmed in various indoor environments, plus an additional `unknown' class containing either background noise or multiple people in the same bounding box. Note that the LIMA dataset is acquired over a significant time period capturing actual people present in a home (\eg residents and `guests'). This makes the dataset interesting as a test bed for long-term analysis, where people's appearance varies significantly, including changes in clothing (see Figure~\ref{fig:appear-changes}). In our experiments, we use a  train-test ratio of $12:1$ implementing a leave-one-session-out approach for cross-validation in order to probe how well performance generalises to different acquisition days. 

%\vspace{-3mm}
\section{Experiments and Results}
\label{sec:exp}
%\vspace{-1mm}

We now perform an extensive system analysis by applying the proposed pipeline mainly to the LIMA dataset. We define as the LIMA baseline the best so-far reported micro precision metric on the dataset achieved by a hybrid M2\&ME approach given in~\cite{LayneLIMA17} -- that is via tracking by recognition-enhanced constrained clustering with multiple enrolment. This approach assigns identities to frames where the accuracy of picking the correct identity as the top-ranking estimate is reported. Against this, we evaluate performance metrics for our approach judging either the performance over all ground truth labels $j$, including the `unknown content' class (\textbf{ALL}), that is $j\in \{0,...,N\}$, or only for known identity ground-truth~(\textbf{p-ID}), that is $j\in \{1,...,N\}$. We use two metrics, \textbf{(i)} \textbf{prec@1 }as the rank-one precision, that is the accuracy of selecting the correct identity for test frames according to the highest class score produced by the final Re-ID CondenseNet $R$; and \textbf{(ii)~mAP} as mean Average Precision  over all considered  classes. Table~\ref{tab:class_res} provides an overview of the results.

\begin{figure}[]
     \centering
     %\vspace{-1mm}
     \includegraphics[width=.99\linewidth]{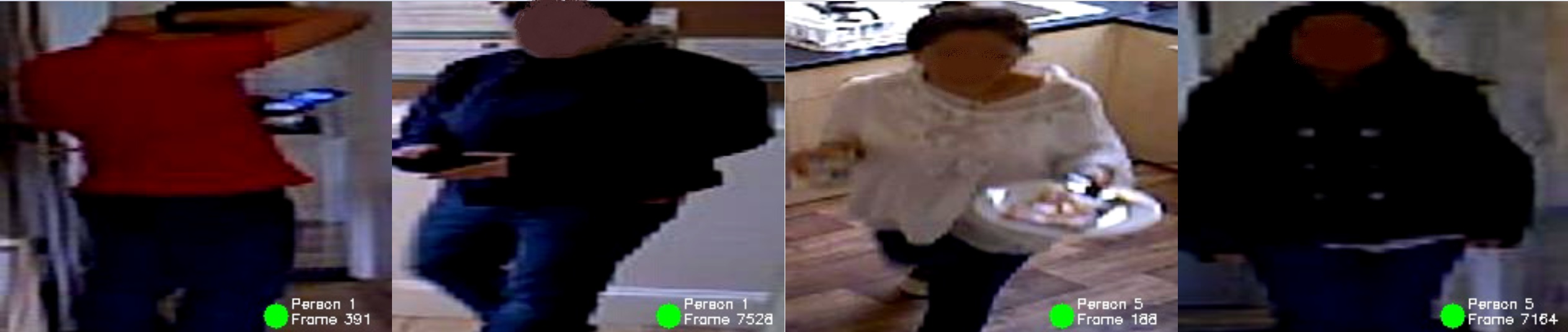}
     %\vspace{-2mm}
     \caption{\small{\textbf{Correct Detections under Changed Appearance.} Examples of two different individual identities at different instances of the same test session (faces have been blurred for privacy reasons)}}
     %\vspace{-3mm}
     \label{fig:appear-changes}
\end{figure}

\begin{figure}[]
     \centering
     %\vspace{-1mm}
     \includegraphics[width=.99\linewidth]{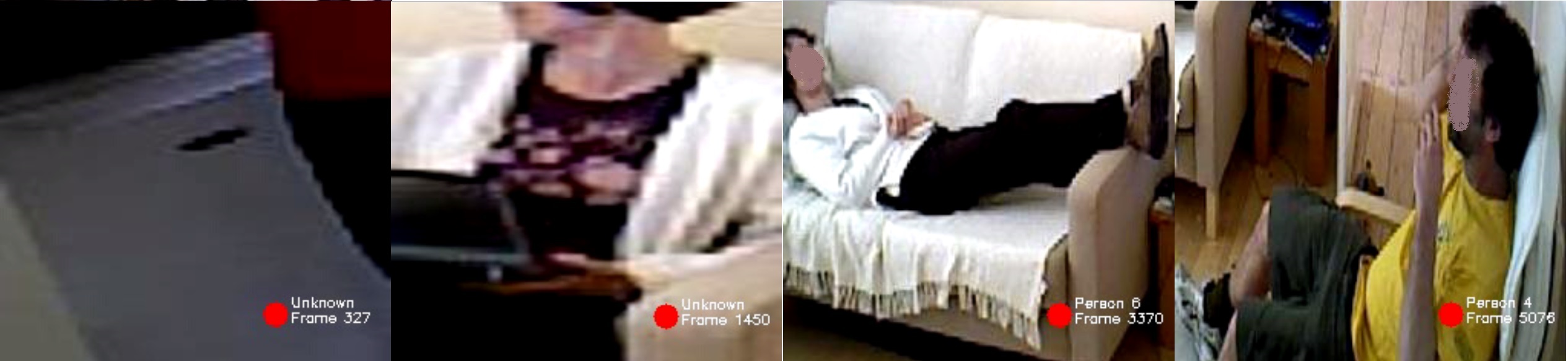}
     %\vspace{-2mm}
     \caption{\small{{\textbf{Failure Cases.} Examples of misidentifications of the `unknown' class and individual identities in challenging positions and without semantic control (faces have been blurred for privacy reasons)}}}
     \vspace{-3mm}
     \label{fig:failures}
\end{figure}

\textbf{Deep CondenseNet without Augmentation ($R$ only) -} The baseline~(Table~\ref{tab:class_res}, row~1) is first compared to results obtained when training CondenseNet ($R$) on original data only~(Table~\ref{tab:class_res}, row~2). This deep compressed network outperforms the baseline \textbf{ALL prec@1} by~$2.88\%$, in particular generalising better for cases of significant appearance change {such as wearing different clothes over the session (\eg. without jacket and wearing a jacket afterwards -- see Figure~\ref{fig:appear-changes})}. The \textbf{p-ID~mAP} results  (\ie discarding the `unknown' class) at $96.28\%$ show that removing destractor content, \ie manual semantic control during the test procedure, can produce scenarios of enhanced performance over filtered test subsets. {Figure~\ref{fig:failures} shows examples of mis-detections for the `unknown' identity and individual identities in absence of the semantic controller.} We will now investigate  how semantic control can be encoded via externally trained networks applied during training.

\textbf{Direct Semantic Control ($FR$) - } Simply introducing a semantic controller $F$ to face-filter the training input of $R$ is, however, counter-productive and reduces performance significantly across all metrics~(Table~\ref{tab:class_res}, row~5). Restricting $R_{}$ to train on only 39\% of the input this way withholds critical identity-relevant information.  % this could be extended for persons as a semantic controller, especially when the resolution become very low

\begin{table}[t]
  \centering\scriptsize
  \caption{\small \textbf{Results for LIMA -} Top rank precision (\textbf{prec@1})  and mean Average Precision (\textbf{mAP}) for baseline (row 1), non-semantically controlled deep CondenseNet approaches (rows 2-4), and various forms of semantic control (rows 5-7). Note  improvements across all metrics when utilising: compressed deep learning (row 2),  augmentation (row 3), and semantically selective  filtering (rows~6-7).}
  \label{tab:class_res}\begin{small}
  \begin{tabular}{|l|l|l|l|l|}
  \hline
  \multicolumn{1}{|c|}{\textbf{No Semantic Control}}     & \multicolumn{1}{c|}{\textbf{ALL prec@1}}    & \multicolumn{1}{c|}{\textbf{p-ID prec@1}}  & \multicolumn{1}{c|}{\textbf{ALL mAP}}  & \multicolumn{1}{c|}{\textbf{p-ID mAP}}            \\ \hline
  1: Baseline (M2\&ME)~\cite{LayneLIMA17}    & \multicolumn{1}{c|}{89.1} & \multicolumn{1}{c|}{-}  & \multicolumn{1}{c|}{-}  & \multicolumn{1}{c|}{-}  \\ \hline
 2: No Augmentation ($R_{}$)        & \multicolumn{1}{c|}{91.98}    & \multicolumn{1}{c|}{93.49}         & \multicolumn{1}{c|}{90.90}    & \multicolumn{1}{c|}{96.28} \\ \hline
  3: Augmentation 24k$G\to R_{}$        & \multicolumn{1}{c|}{\textit{92.43}}  & \multicolumn{1}{c|}{\textit{94.27}}  & \multicolumn{1}{c|}{\textit{91}}      & \multicolumn{1}{c|}{\textit{96.95}}  \\ \hline
  4: Augmentation 48k$G\to R_{}$        &  \multicolumn{1}{c|}{91.74}     & \multicolumn{1}{c|}{93.48} &  \multicolumn{1}{c|}{90.61} &  \multicolumn{1}{c|}{96.54}      \\ \hline
 \hline
  \multicolumn{1}{|c|}{\textbf{Semantic Control via $F$}}     & \multicolumn{1}{c|}{\textbf{ALL prec@1}}    & \multicolumn{1}{c|}{\textbf{p-ID prec@1}}  & \multicolumn{1}{c|}{\textbf{ALL mAP}}  & \multicolumn{1}{c|}{\textbf{p-ID mAP}}            \\
\hline
  5: No Augmentation ($FR_{}$)      &   \multicolumn{1}{c|}{82.02}        &  \multicolumn{1}{c|}{92.14}         &  \multicolumn{1}{c|}{72.90}   &  \multicolumn{1}{c|}{95.48}           \\ \hline
 6: Augmentation $F$322k$G\to R_{}$ &    \multicolumn{1}{c|}{\textbf{92.58}}                          &    \multicolumn{1}{c|}{\textbf{94.57}}       &    \multicolumn{1}{c|}{\textbf{91.14}}      &    \multicolumn{1}{c|}{97.02}        \\ \hline
 7: $($24k$G_0$+$F$24k$G_j)\to R_{}$ &    \multicolumn{1}{c|}{92.44}      &    \multicolumn{1}{c|}{94.37}       &    \multicolumn{1}{c|}{90.96}         &    \multicolumn{1}{c|}{\textbf{97.04}}                        \\ \hline
  \end{tabular}\end{small}
%\vspace{-10pt}
\end{table}

\begin{figure}[b]
     \centering
     %\vspace{-1mm}
     \includegraphics[width=\linewidth]{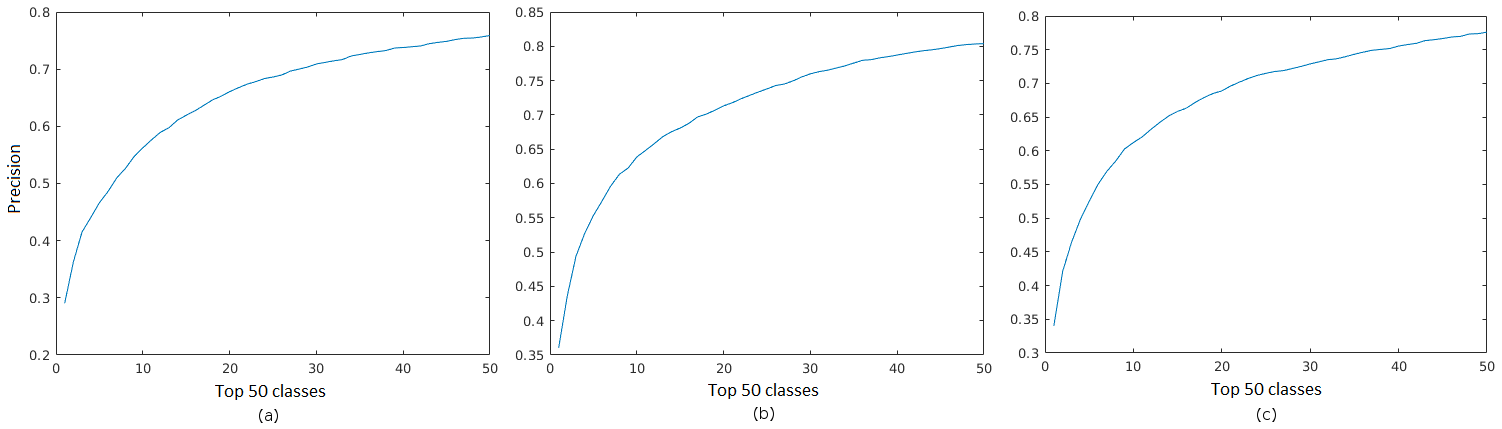}
     %\vspace{-6mm}
     \caption{\small {\textbf{CMC Curves for DukeMTMC-reID.} Visualisation of the precision over the top-$N$ classes (where $N\leq50$) for various experimental settings detailed in Table~\ref{tab:class_match} over rows 3-5:   \textbf{(a)} row 3: training without augmentation~($R$), \textbf{(b)} row 4: basic DCGAN augmentation (24k$G\to R$), \textbf{(c)} row 5: transfer augmentation via synthesis based on different dataset Market1501 24k(Market1501)$G\to R$.}}
     %\vspace{-2mm}
     \label{fig:CMC-curves}
\end{figure}

{\textbf{Augmentation via {DC}GANs ($G$) - } Instead of restricting training input to the Re-ID network~$R$, we therefore analyse how Re-ID performance is affected when semantic control is applied to generic {DC}GAN-synthesis via $G$ of a cross-identity person class as suggested in~\cite{zheng2017unlabeled}. {Figure~\ref{fig:DCGAN_samples} shows   examples of generated images and how the semantic controller~affects to the synthesis appearance.} Augmentation of training data with $24$k synthesised samples without semantic control~(Table~\ref{tab:class_res}, row {3}) improves performance slightly across all metrics, confirming benefits discussed in more detail in~\cite{zheng2017unlabeled}. Table~\ref{tab:class_match} confirms that applying such {DC}GAN synthesis together with CondenseNet compression to the DukeMTMC-reID dataset produce results comparable to~\cite{BowKissme15}. Figure~\ref{fig:CMC-curves} provides further details on these experimental outcomes. Note that whilst the large deep ResNet50+LSRO~\cite{zheng2017unlabeled} approach outperforms our compressed network significantly (Table~\ref{tab:class_match}, row 6), this comes at a cost of increasing the parameter cadinality by about an order of magnitude\footnote{It requires approximately 8$\times$ fewer parameters and operations to achieve a comparable accuracy \wrt other dense nets (\ie around 600 million less operations to perform inference on a single image)~\cite{huang2017condensenet}}.
Moreover, non-controlled synthesis is generally limited. Indeed, on LIMA no further improvements can be made by scaling up synthesis beyond $24$k, indeed performance drops slightly across all metrics and overfitting to the synthetised~data can be observed (Table~\ref{tab:class_res}, row~{4}). We now introduce semantic control to the input of augmentation and observe that the scaling-up limit can be  lifted. Diminishing returns take over at levels above $300$k though {(\ie 54\% of synthesis \wrt original training data)}. We report results when synthesising $322$k of imagery via $G$ improving results for all metrics (Table~\ref{tab:class_res}, row 6). We note that these improvements are achieved by synthesising distractors rather than provision of individual-specific augmentations.

\begin{table}[t]
  \centering\scriptsize
  \caption{\small \textbf{Results for DukeMTMC-reID -} Top rank precision (\textbf{prec@1}) for classification and Single-Query (S-Q) performance. Our results outperform~\cite{BowKissme15} when using augmentation (row 4), or using Market1501 as synthesis input (row 5). However, the performance of the $8\times$ larger ResNet50+LSRO~\cite{zheng2017unlabeled} cannot be achieved in our setting of compression for lightweight deployment.}
  \label{tab:class_match}\begin{small}
    \begin{tabular}{|l|l|l|l|l|l|}
    \hline
    \multicolumn{1}{|c|}{\textbf{Method / No Semantic Control}}         & \multicolumn{1}{c|}{\textbf{prec@1}}      & \multicolumn{1}{c|}{\textbf{prec@5}}  & \multicolumn{1}{c|}{\textbf{mAP}}  \vline\vline\vline\vline\vline    & \multicolumn{1}{c|}{\textbf{CMC@1 S-Q}}        & \multicolumn{1}{c|}{\textbf{mAP S-Q}}            \\ \hline
    1: Baseline BoW + KISSME~\cite{BowKissme15}         &   \multicolumn{1}{c|}{-}         &  \multicolumn{1}{c|}{-}      &  \multicolumn{1}{c|}{-}      \vline\vline\vline\vline\vline  &  \multicolumn{1}{c|}{25.13}      &  \multicolumn{1}{c|}{12.17}               \\ \hline
    2: Baseline LOMO + XQDA~\cite{BowKissme15}  &   \multicolumn{1}{c|}{-}         &  \multicolumn{1}{c|}{-}    &  \multicolumn{1}{c|}{-}    \vline\vline\vline\vline\vline    &  \multicolumn{1}{c|}{30.75}      &  \multicolumn{1}{c|}{17.04}             \\ \hline
    3: No Augmentation ($R$)    & \multicolumn{1}{c|}{87.70}      & \multicolumn{1}{c|}{95.54}  & \multicolumn{1}{c|}{87.79}  \vline\vline\vline\vline\vline   & \multicolumn{1}{c|}{\textit{29.04}}        & \multicolumn{1}{c|}{\textit{15.99}}            \\ \hline
    4: Augmentation 24k$G\to R$         & \multicolumn{1}{c|}{88.08}      & \multicolumn{1}{c|}{95.73}  & \multicolumn{1}{c|}{88.26}  \vline\vline\vline\vline\vline         & \multicolumn{1}{c|}{\textbf{36.45}}        & \multicolumn{1}{c|}{\textbf{21.11}}            \\ \hline
    5: Transfer 24k(\begin{footnotesize}Market1501\end{footnotesize})$G\to R$         & \multicolumn{1}{c|}{\textbf{88.84}}      & \multicolumn{1}{c|}{\textbf{95.82}}  & \multicolumn{1}{c|}{\textbf{88.64}}  \vline\vline\vline\vline\vline         & \multicolumn{1}{c|}{\textit{35.95}}        & \multicolumn{1}{c|}{\textit{20.6}}    \\ \hline\hline
    6: ResNet50+LSRO~\cite{zheng2017unlabeled} \begin{footnotesize}(8x larger)\end{footnotesize}      &   \multicolumn{1}{c|}{-}         &  \multicolumn{1}{c|}{-}    &  \multicolumn{1}{c|}{-}    \vline\vline\vline\vline\vline    &  \multicolumn{1}{c|}{67.68}      &  \multicolumn{1}{c|}{47.13}             \\ \hline
    \end{tabular}\end{small}%\vspace{-6pt}
\end{table}

\begin{figure}[t]
     \centering
     %\vspace{-1mm}
     \includegraphics[width=\linewidth]{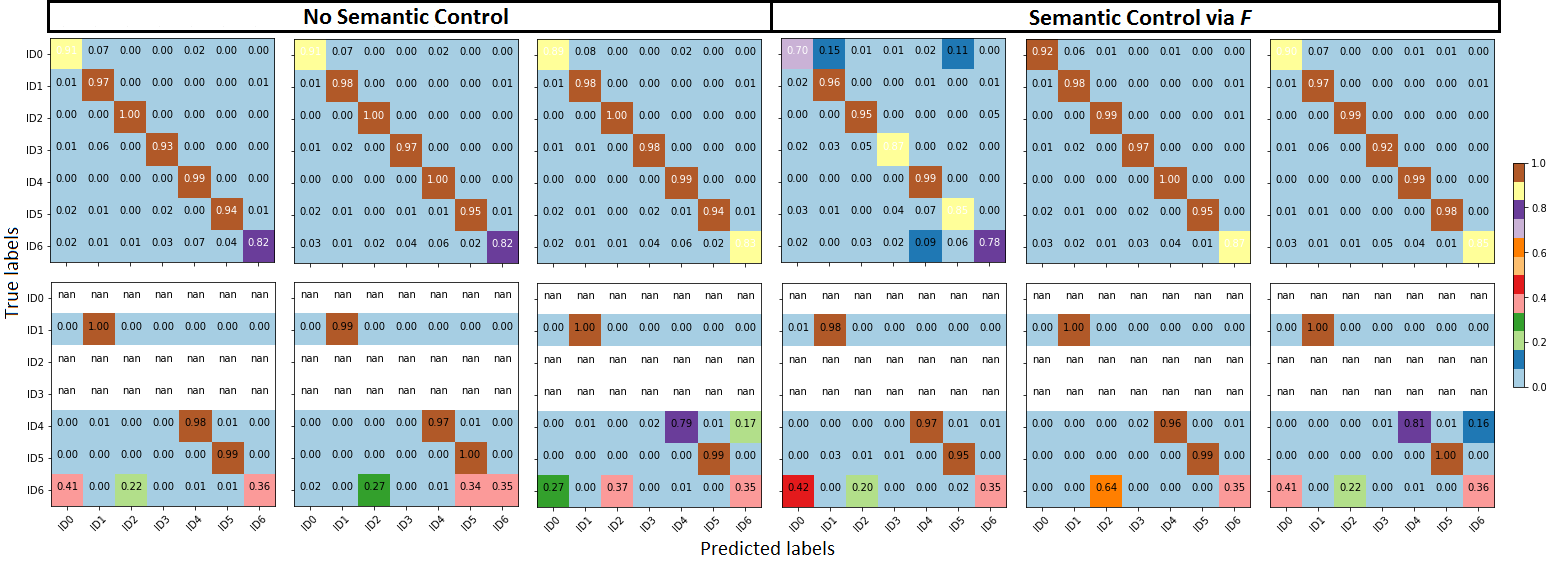}
     %\vspace{-6mm}
     \caption{\small \textbf{Some Results as Confusion Matrices.} {Columns from left to right correspond to the experimental settings grouped by the presence of semantic selection, according to Table~\ref{tab:class_res} rows 2-4 and 5-7, respectively. Top and bottom rows correspond to two challenging test sessions from the LIMA dataset, where some IDs may not be present (nan true-label values).}}
     %\vspace{-3mm}
     \label{fig:cm}
\end{figure}

\textbf{Individual-specific Augmentation ($G_0$ + $G_j$) - } To explore class-specific augmentation we train an entire set of {DC}GANs, \ie produce generators $G_j$ and $G_0$, respectively as specific identity and non-identity synthesis networks, and apply semantic control  $F$ to the identity classes $j\in \{1,...,N\}$. We observe that when balancing the synthesis of training imagery across all classes equally only slightly improves on  \textbf{p-ID~mAP}, whilst other measures cannot be advanced (Table~\ref{tab:class_res}, row 7). Figure~\ref{fig:cm} provides further result visualisations. The limited improvements of this approach compared to non-identity-specific training (despite  synthesis of overall more training data) suggest that, for the LIMA setup at least, person individuality can indeed be encoded by augmentation-supported modelling of a large, generic `person' class against a more limited, non-augmented  representation of individuals. {Furthermore, experiments on the most challenging LIMA sessions demonstrate that the pre-trained generator $G$ can generalize at re-training individual-specific generators $G_0$ and $G_j$ so as to reduce training cost of DCGAN indvidual-specific augmentation (\eg see identity $j=1$~in~Figure~\ref{fig:dcgan-training}).}

%\vspace{-1mm}
\section{Conclusion}
\label{sec:conclusion}
%\vspace{-1mm}
%\textcolor{green}{
We introduced a deep person {Re-ID} approach that brought together semantically selective data augmentation with clustering-based network compression to produce light and fast inference networks. In particular, we showed that augmentation via sampling from a DCGAN, whose discriminator is constrained by a semantic face detector, can outperform the state-of-the-art on the LIMA dataset for long-term monitoring in indoor living environments. To explore the applicability of our framework without face detection in outdoor scenarios, we also considered well-known datasets for person {Re-ID} aimed at people matching, achieving competitive performance on the DukeMTMC-reID dataset. 

{Exploring generic and effective semantic controllers as part of discriminator networks is an immediate extension of our work, specially to deal with low resolution images, as well as learning generators from other person-like representations broadly across the Re-ID domain.}

\bibliography{bmvc2018bib}
\end{document}